\documentclass[10pt, a4paper, journal, two side]{ICTR}

\usepackage{graphicx}
\usepackage[table]{xcolor}
\usepackage{epstopdf} 
\usepackage{subfigure}

\usepackage[ruled,linesnumbered]{algorithm2e}

\usepackage{multirow}
\usepackage{tabularx}
\usepackage{makecell}
\usepackage{booktabs}
\usepackage{arydshln}

\usepackage{textcomp}
\usepackage{cite}
\usepackage[square, comma, numbers, sort&compress]{natbib}

\usepackage{listings} 
\usepackage{lineno}
\usepackage{balance}

\usepackage{xspace}  % \xspace
\usepackage{xcolor}  % \color{}
\usepackage{colortbl}
\usepackage[table]{xcolor}
\usepackage{multirow}  % \multirow
\usepackage[
    pdftitle={Leveraging Model Soups to Classify Intangible Cultural Heritage Images from the Mekong Delta},
    pdfauthor={Quoc-Khang and Minh-Thien et al.},
    pdfkeywords={ICH Classification, Model Soups, CoAtNet, Hybrid Model, Ensemble Learning}
]{hyperref} % for links, metadata and bookmarks
\usepackage[T5]{fontenc}  % Encoding hỗ trợ tiếng Việt
\usepackage[utf8]{inputenc}  % Nếu bạn gõ Unicode
\usepackage{booktabs}  % \toprule, \midrule, etc
\usepackage{makecell}  % \makecell used in tables
\usepackage{amsmath}
\usepackage{amssymb}

\usepackage{newtxmath}
\usepackage{amsfonts}
\usepackage[revision]{thirdparty/revdiff} %  LaTeX revision and diff package, choose mode from [revision, clean]

\newcommand{\modelsoups}{\textit{model soups}\xspace}

\begin{document}
% Huu-Hoa Nguyen$^1$
\title{Leveraging Model Soups to Classify\\Intangible Cultural Heritage Images\\from the Mekong Delta}
\AuthorName{Quoc-Khang Tran$^1$, Minh-Thien Nguyen$^1$, Nguyen-Khang Pham$^2$}
\AuthorAddress  {$^1$ College of Information and Communication Technology, Can Tho University \\ $^2$ School of Graduate, Can Tho University}
\Correspondence {Nguyen-Khang Pham, pnkhang@ctu.edu.vn}

%% ICTR to add 
\DateReceived{28 Aug 2025}
\DateRevised {}
\DateAccepted{24 Sep 2025}
\DateEarlyAccess{1 Oct 2025}
\DOI{10.32913/mic-ict-research.v2025.n3.1395}
\Volume{2025}
\Number{3}
\Month{November}
\ICTRleftmark{\ICTRJournal}
\ICTRrightmark

\maketitle

\setcounter{page}{1}

\begin{abstract}
The classification of Intangible Cultural Heritage (ICH) images in the Mekong Delta poses unique challenges due to limited annotated data, high visual similarity among classes, and domain heterogeneity. In such low-resource settings, conventional deep learning models often suffer from high variance or overfit to spurious correlations, leading to poor generalization. To address these limitations, we propose a robust framework that integrates the hybrid CoAtNet architecture with \modelsoups, a lightweight weight-space ensembling technique that averages checkpoints from a single training trajectory -- \textit{without increasing inference cost}. CoAtNet captures both local and global patterns through stage-wise fusion of convolution and self-attention. We apply two ensembling strategies -- \textit{greedy} and \textit{uniform} soup -- to selectively combine diverse checkpoints into a final model. Beyond performance improvements, we analyze the ensembling effect through the lens of bias–variance decomposition. Our findings show that \modelsoups reduces variance by stabilizing predictions across diverse model snapshots, while introducing minimal additional bias. Furthermore, using cross-entropy-based distance metrics and Multidimensional Scaling (MDS), we show that \modelsoups selects geometrically diverse checkpoints, unlike Soft Voting, which blends redundant models centered in output space. Evaluated on the ICH-17 dataset (7,406 images across 17 classes), our approach achieves \textbf{state-of-the-art} results with 72.36\% top-1 accuracy and 69.28\% macro F1-score, outperforming strong baselines including ResNet-50, DenseNet-121, and ViT. These results underscore that diversity-aware checkpoint averaging provides a principled and efficient way to reduce variance and enhance generalization in culturally rich, data-scarce classification tasks.
\end{abstract}


% \section{Introduction}

% Introduce the problem, its meaning, role and challenge in practical applications. Present relevant studies, pointing out the remaining problems. Indicate the research objectives and the results to be achieved. Outline the structure of the rest of the paper.
 
% \section{Research Methodology}
% The section presents the main research contents, which can be extended to the sections before and after this section.

% \section{Experiment Results}

% Presenting theoretical experimental results in a practical environment, Table \ref{tab2} and Figure \ref{fig4} are presented accurately, clearly and sharply.

% \begin{figure}[!h]

% \centering
% \includegraphics[scale=0.3]{images/iso27005process3.png}
% \caption{E-UTRAN RACH Setup attempts - predicted data versus actual data} \label{fig4}
% \end{figure}

% \begin{table}[!ht]
% \centering
% \caption{Error rates of each prediction.}\label{tab2}
% \begin{tabular}{|c|c|}
% \hline
% \textbf{Behaviors} &  \textbf{MAPE (\%)}\\
% \hline
% \hline
% E-UTRAN RACH Setup Attempts &  27\\
% \hline
% RRC Setup Attempts & 10\\
% \hline
% E-UTRAN Data Radio Bearer Setup Attempts & 12\\
% \hline
% \end{tabular}
% \end{table}

% \section{Conclusion}
% Briefly present the results achieved, the remaining issues and the future development direction of the paper

% \small{
% \section*{Acknowledgement}
% Type here

\begin{figure}[!ht]
  \centering
  \includegraphics[width=\linewidth]{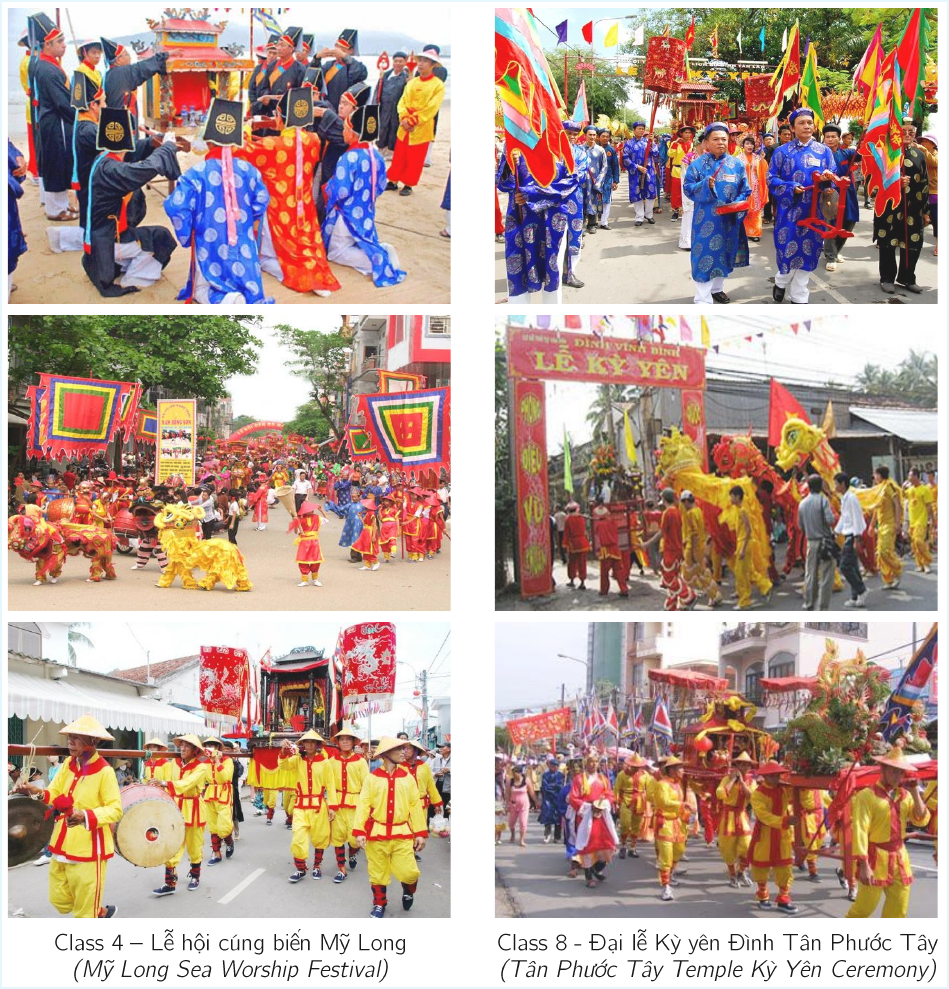}
  % \caption{Similar context between class 4 (left) and class 8 (right)}
  \caption{An illustrative example of the challenging nature of this task: class 4 (left) and class 8 (right) share highly similar visual contexts, making them difficult to distinguish.}
  \label{fig:class-4-and-8-comparison}
\end{figure}

\section{Introduction}

The classification of Intangible Cultural Heritage (ICH) images in the Mekong Delta plays a crucial role in cultural preservation, documentation, and digital dissemination. However, building accurate image classifiers for ICH is inherently challenging due to the diversity of cultural practices, subtle inter-class variations, and the limited scale of high-quality annotated data.
% \newpage
To address these issues, we propose a novel classification framework based on the CoAtNet architecture and model ensembling via \modelsoups~\cite{wortsman2022modelsoupsaveragingweights}. CoAtNet is a hybrid vision model that combines convolutional operations with self-attention mechanisms, effectively capturing both local patterns and long-range dependencies in images~\cite{dai2021coatnetmarryingconvolutionattention}. Its flexibility and representational power make it a suitable backbone for our task.

Rather than training multiple independent models with different seeds or configurations, we generate an ensemble by collecting multiple checkpoints from a single CoAtNet training process. These checkpoints, representing different stages of model optimization, are aggregated through the \modelsoups\ technique --- a method that averages the weights of multiple checkpoints without requiring additional joint training~\cite{wortsman2022modelsoupsaveragingweights}. This simple yet effective strategy enhances generalization by interpolating between distinct solutions in the weight space, thereby reducing overfitting and leveraging the complementary strengths of intermediate model states.

Our experimental results on a curated 17-class ICH dataset from the Mekong Delta show that the proposed framework significantly outperforms standalone fine-tuned models, highlighting its potential as a practical and scalable solution for heritage-related image classification.
\section{Related Work}
\label{sec:related}

The classification of intangible cultural heritage (ICH) imagery has received increasing attention with the advancement of deep learning techniques in cultural informatics. In early studies, conventional convolutional neural networks (CNNs) and traditional machine learning approaches were employed to recognize ICH categories in Vietnam. Notably, Do et al.~\cite{do2021visual} introduced a 17-class ICH image dataset from the Mekong Delta and benchmarked various feature extractors such as VGG19, ResNet50, Inception-v3, and Xception in combination with support vector machines (SVMs), attaining an accuracy of 65.32\%. Subsequently, Tran et al.~\cite{tran2024fusing} improved upon this result by fusing deep features and classifier outputs through logistic regression, achieving 66.76\% accuracy. Despite these efforts, model performance remained modest, and little attention was paid to ensemble strategies or architectural advancements.

Ensemble learning has long been recognized as an effective strategy for improving generalization in classification tasks. A recent and notable contribution is \modelsoups~\cite{wortsman2022modelsoupsaveragingweights}, a method that combines the weights of independently fine-tuned models to produce a single, more robust model without requiring additional inference time. This approach leverages the diversity of multiple checkpoints of a model to create a solution that generalizes better than any single model alone.

At the architectural level, hybrid models have emerged as strong alternatives to traditional CNNs. Among them, CoAtNet~\cite{dai2021coatnetmarryingconvolutionattention} integrates convolutional operations with self-attention mechanisms, offering a unified design that balances local feature extraction with global context modeling. CoAtNet has demonstrated state-of-the-art performance across a wide range of vision benchmarks and is particularly well-suited for tasks involving limited annotated data, such as ICH image classification.

In this work, we build upon these recent advances by applying the \modelsoups technique to an ensemble of CoAtNet models for multi-class ICH classification. To the best of our knowledge, this is the first study that leverages CoAtNet-based \modelsoups to improve classification performance on cultural heritage datasets.
\section{Research Methodology}

\subsection{Classification of Intangible Cultural Heritage Images}

The Mekong Delta is renowned for its rich repository of Intangible Cultural Heritage (ICH), encompassing a diverse array of cultural expressions. These include traditional practices such as \textit{Đờn ca Tài tử Nam Bộ (Art of Đờn ca tài tử music and song in southern Vietnam)} and \textit{Văn hoá Chợ nổi Cái Răng (Cái Răng Floating Market Culture)}, traditional handicrafts like  \textit{Nghề dệt chiếu (Mat weaving)} and \textit{Nghề đan tre (Bamboo weaving)}, as well as traditional festivals, including \textit{Lễ hội Ok Om Bok của người Khmer (Khmer Ok Om Bok festival)} and \textit{Lễ hội Vía Bà Chúa Xứ Núi Sam (Festival of Bà Chúa Xứ Goddess at Sam Mountain)}. These intangible cultural heritages play a pivotal role in preserving Vietnam’s cultural values and enhancing the diversity of intangible cultural heritage in the Mekong Delta region.

In this study, we focus on the classification of images of intangible cultural heritage. Specifically, our goal is to improve classification performance on a dataset comprising images from 17 ICH categories (referred to as the ICH-17 dataset). Do et al.~\cite{do2021visual} constructed this dataset by collecting images via Google search using text-based queries, followed by manual post-processing to remove noisy or irrelevant images. Detailed information about the dataset is provided in Table~\ref{tab:ich-17-dataset}.

Despite manual post-processing, the ICH-17 dataset still contains a notable proportion of noisy, irrelevant images . Additionally, the images exhibit significant contextual diversity, and certain categories share relatively similar contexts, such as \textit{Lễ hội cúng biển Mỹ Long (Mỹ Long Sea Worship Festival}) (class 4) and \textit{Đại lễ Kỳ yên Đình Tân Phước Tây (Tân Phước Tây Temple Kỳ Yên Ceremony}) (class 8) (see Fig.~\ref{fig:class-4-and-8-comparison}). These characteristics of the ICH-17 dataset pose significant challenges to achieving high classification performance, as evidenced in prior studies~\cite{do2021visual,tran2024fusing,10.1007/978-3-030-38364-0_17}.

\subsection{CoAtNet: Hybrid Convolution-Attention Design}

\textbf{CoAtNet}~\cite{dai2021coatnetmarryingconvolutionattention} is adopted in this study as a hybrid neural architecture unifying convolutional operations with self-attention in a stage-wise design to capitalize on the strengths of both paradigms. Specifically, the network consists of five stages, where the initial stage ($S_0$) is a convolutional stem composed of standard convolutional layers, while $S_1$ and $S_2$ employ MBConv blocks~\cite{sandler2019mobilenetv2invertedresidualslinear} with depthwise separable convolutions and squeeze-and-excitation mechanisms~\cite{hu2019squeezeandexcitationnetworks} to efficiently capture local features. The later stages ($S_3$ and $S_4$) transition to Transformer blocks~\cite{vaswani2023attentionneed} incorporating relative self-attention and convolutional feed-forward networks to model global dependencies. This C--C--T--T configuration enables CoAtNet to maintain strong inductive biases for spatial generalization while enhancing its capacity to learn long-range interactions.  
% Comparative experiments are conducted against several baselines: \textbf{ResNet-50}~\cite{he2015deepresiduallearningimage}, a canonical CNN with residual connections that facilitate deep learning; \textbf{DenseNet-121}~\cite{huang2018denselyconnectedconvolutionalnetworks}, which improves information flow via dense inter-layer connectivity; and \textbf{Vision Transformer (ViT)}~\cite{dosovitskiy2021imageworth16x16words}, a fully attention-based model that directly processes image patches and lacks convolutional priors to evaluate performance in ICH classification.
% To evaluate the effectiveness of this architecture in the classification of intangible cultural heritage imagery, we conduct comparative experiments against representative baselines: \textbf{ResNet-50}~\cite{he2015deepresiduallearningimage}, a canonical CNN with residual connections that facilitate deep learning; \textbf{DenseNet-121}~\cite{huang2018denselyconnectedconvolutionalnetworks}, which improves information flow via dense inter-layer connectivity; and \textbf{Vision Transformer (ViT)}~\cite{dosovitskiy2021imageworth16x16words}, a fully attention-based model that directly processes image patches and lacks convolutional priors.

% This evaluation framework allows for a thorough investigation of how architectural choices influence performance in the ICH classification task.

Comparative experiments were conducted against three representative baselines: \textbf{ResNet-50}~\cite{he2015deepresiduallearningimage}, a classic CNN architecture featuring residual connections that facilitate effective deep learning; \textbf{DenseNet-121}~\cite{huang2018denselyconnectedconvolutionalnetworks}, which enhances information flow through dense, inter-layer connectivity; and \textbf{Vision Transformer (ViT)}~\cite{dosovitskiy2021imageworth16x16words}, a purely attention-based model that processes image patches directly and operates without convolutional priors. These comparisons assess the architecture’s performance in intangible cultural heritage classification.

\begin{figure*}[ht]
  \centering
  \includegraphics[width=\linewidth]{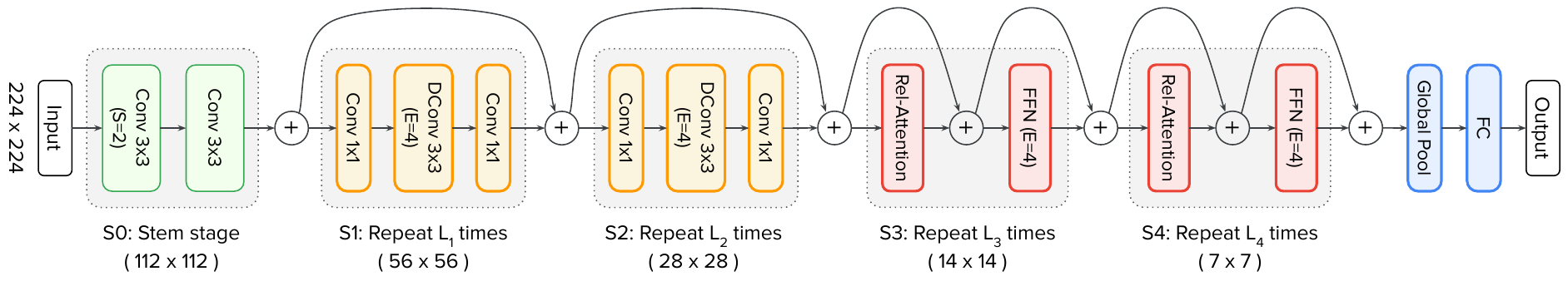}
  \caption{Overview of CoAtNet architecture. The model comprises five stages, gradually transitioning from convolutional blocks (MBConv) to Transformer blocks (self-attention). Each stage reduces spatial resolution while increasing the number of channels. Image credit:~\cite{dai2021coatnetmarryingconvolutionattention}.}
  \label{fig:coatnet}
\end{figure*}

\begin{table}[!t]
    \centering
    \small
    \caption{Details for the ICH-17 dataset}
    \resizebox{\linewidth}{!}{%
    \begin{tabular}{c l c} \toprule
        \textbf{No.} & \textbf{Category} & \textbf{\texttt{\#} Images} \\
        \midrule
        1 & \makecell[l]{Đờn ca Tài tử Nam Bộ \\ \textit{Art of Don ca Tai tu music and song in southern Viet Nam}} & 513 \\
        \midrule
        2 & \makecell[l]{Nghệ thuật Chầm riêng chà pây của người Khmer \\ \textit{Khmer Cham rieng cha pay art}} & 185 \\
        \midrule
        3 & \makecell[l]{Nghề dệt chiếu \\ \textit{Mat weaving}} & 642 \\
        \midrule
        4 & \makecell[l]{Lễ hội cúng biển Mỹ Long \\ \textit{My Long Sea worship festival}} & 398 \\
        \midrule
        5 & \makecell[l]{Nghệ thuật sân khấu Dù Kê của người Khmer \\ \textit{Du Ke (or Lakhon Bassac) Theater Arts of the Khmer people}} & 404 \\
        \midrule
        6 & \makecell[l]{Lễ hội Ok Om Bok của người Khmer \\ \textit{Khmer Ok Om Bok festival}} & 465 \\
        \midrule
        7 & \makecell[l]{Lễ hội Vía Bà Chúa Xứ núi Sam \\ \textit{Festival of Ba Chua Xu Goddess at Sam Mountain}} & 405 \\
        \midrule
        8 & \makecell[l]{Đại lễ Kỳ yên Đình Tân Phước Tây \\ \textit{Tan Phuoc Tay temple Ky Yen ceremony}} & 223 \\
        \midrule
        9 & \makecell[l]{Lễ hội vía Bà Ngũ hành \\ \textit{Festival of the Five Elements Goddess}} & 569 \\
        \midrule
        10 & \makecell[l]{Lễ làm chay \\ \textit{Tam Vu's Alms Giving festival/Tam Vu's Making Offerings festival}} & 365 \\
        \midrule
        11 & \makecell[l]{Nghề đóng xuồng ghe Long định \\ \textit{The craft of building wooden boats in Long Dinh}} & 281 \\
        \midrule
        12 & \makecell[l]{Nghề đan tre \\ \textit{Bamboo weaving}} & 639 \\
        \midrule
        13 & \makecell[l]{Lễ cúng Việc \\ \textit{Ancestor worship ceremony/Ancestral ritual}} & 447 \\
        \midrule
        14 & \makecell[l]{Lễ hội Đua bò Bảy Núi \\ \textit{Bay Nui bull racing festival}} & 449 \\
        \midrule
        15 & \makecell[l]{Lễ hội Nghinh Ông \\ \textit{Nghinh Ong festival/Whale worship festival}} & 522 \\
        \midrule
        16 & \makecell[l]{Lễ hội anh hùng Trương \\ \textit{The Death Anniversary of National Hero Truong Dinh}} & 361 \\
        \midrule
        17 & \makecell[l]{Văn hóa Chợ nổi Cái Răng \\ \textit{Cai Rang floating market culture}} & 538 \\
        \midrule
        & Total images & 7406
    \end{tabular}
    \label{tab:ich-17-dataset}
    }%
\end{table}

\subsection{Weight-Space Ensembling via Model Soups}
\label{subsec:model-soups}

% To enhance model generalization without increasing inference cost, we adopt the \modelsoups technique~\cite{wortsman2022modelsoupsaveragingweights}, which constructs a single model by averaging the weights of multiple checkpoints in parameter space. Unlike conventional ensembles that aggregate predictions at inference time, \modelsoups produces a single model with no additional runtime overhead.

% Our ensembling strategy is grounded in the \textit{greedy soup} procedure, which incrementally selects a subset of high-performing checkpoints along the CoAtNet training trajectory and averages their weights to form the final model. The process begins with the checkpoint that achieves the highest validation accuracy. Subsequent candidates are included only if their addition improves the validation performance of the averaged model.

The \modelsoups technique~\cite{wortsman2022modelsoupsaveragingweights} is adopted to enhance generalization without increasing inference cost. A greedy soup procedure is employed: the checkpoint achieving the highest validation accuracy is selected first, and additional candidates are incorporated only if they yield improved validation performance.

Let \( \{\theta^{(1)}, \theta^{(2)}, \ldots, \theta^{(T)}\} \) denote the set of saved checkpoints at different epochs. The greedy selection identifies a subset \( \mathcal{S} \subseteq \{1, \dots, T\} \) of checkpoint indices. The final ensemble model, referred to as a \textit{uniform soup}, is then computed as:

\begin{equation}
\theta_{\text{soup}} = \frac{1}{|\mathcal{S}|} \sum_{k \in \mathcal{S}} \theta^{(k)},
\end{equation}

where \( \theta^{(k)} \) denotes the parameter vector of the \(k\)-th selected checkpoint. By integrating greedy selection with uniform averaging, the method filters out suboptimal checkpoints while retaining the simplicity and generalization benefits of weight-space ensembling~\cite{wortsman2022modelsoupsaveragingweights,vaswani2023attentionneed}. The detailed procedure is outlined in Algorithm~\ref{alg:greedy-soup}.

\vspace{1em}
\begin{algorithm}[ht]
\caption{Greedy Soup Selection Algorithm}
\label{alg:greedy-soup}
\KwIn{
  Set of candidate checkpoints $\{\theta^{(1)}, \theta^{(2)}, \dots, \theta^{(T)}\}$\;
  Validation set $\mathcal{D}_{\text{val}}$\;
}
\KwOut{
  Selected index set $\mathcal{S}$ and final soup weights $\theta_{\text{soup}}$
}
Initialize $\mathcal{S} \leftarrow \{\arg\max_{k} \text{Acc}(\theta^{(k)}; \mathcal{D}_{\text{val}})\}$ \tcp*{Start with best checkpoint}

\For{$k = 1$ \KwTo $T$}{
  \If{$k \notin \mathcal{S}$}{
    Compute temporary average: $\theta_{\text{temp}} = \frac{1}{|\mathcal{S}| + 1} \left( \sum_{j \in \mathcal{S}} \theta^{(j)} + \theta^{(k)} \right)$\;
    Evaluate temporary accuracy: $\text{Acc}_{\text{temp}} = \text{Accuracy}(\theta_{\text{temp}}; \mathcal{D}_{\text{val}})$\;

    \If{$\text{Acc}_{\text{temp}} \geq \text{Acc}\left( \frac{1}{|\mathcal{S}|} \sum_{j \in \mathcal{S}} \theta^{(j)} \right)$}{
      $\mathcal{S} \leftarrow \mathcal{S} \cup \{k\}$ \tcp*{Accept checkpoint}
    }
  }
}
Compute final average: $\theta_{\text{soup}} = \frac{1}{|\mathcal{S}|} \sum_{k \in \mathcal{S}} \theta^{(k)}$\;
\Return{$\mathcal{S}, \theta_{\text{soup}}$}
\end{algorithm}

Empirically, we observed that this greedy-then-uniform ensembling approach consistently outperforms any individual checkpoint, offering improved robustness and better generalization -- particularly in the context of multi-class classification for intangible cultural heritage images.
% \newpage
\section{Experiments}

\subsection{Dataset}

We conducted experiments on the ICH-17 dataset, which consists of 17 different categories with a total of 7,406 images. Specific details for each class are presented in Table~\ref{tab:image_counts}. The dataset initially consisted of two sets: a training set (6,657 images) and a test set (749 images) (this test set is the same as the one used in the previous study~\cite{do2021visual}).
% We randomly split the original training set into a new training set (6,057 images) and a validation set (600 images). The new training set was used for training the models, the validation set was used to evaluate the model's performance during training as well as for \modelsoups, and the test set was used solely for final result evaluation.
The original training set was randomly partitioned into a new training subset comprising 6,057 images and a validation subset containing 600 images. The training subset was utilized for model training, the validation subset was employed to monitor performance during training and to construct \modelsoups, and the test set was reserved exclusively for final evaluation.

% We conducted experiments on the ICH-17 dataset, which consists of 17 different categories with a total of 7,406 images. Specific details for each class are presented in Table~\ref{tab:image_counts}. For ease of comparison, similar to previous study~\cite{do2021visual,tran2024fusing,10.1007/978-3-030-38364-0_17}, we randomly split the dataset into a training set (6,057 images), a validation set (600 images), and a \textbf{test set} (741 images). The training set was used for training the models, the validation set was used to evaluate the model's performance during training as well as for \modelsoups, and the test set was used solely for final result evaluation.

\begin{table}[htbp]
\centering
\caption{Number of images per class in the training, validation, and test sets of the ICH-17 dataset}
\label{tab:image_counts}
\begin{tabular}{c|c|c|c|c}
\textbf{Class} & \textbf{Train} & \textbf{Validation} & \textbf{Test} & \textbf{Total} \\
\hline
% 1  & 416 & 46  & 51  & 513 \\
% 2  & 150 & 17  & 18  & 185 \\
% 3  & 520 & 58  & 64  & 642 \\
% 4  & 322 & 36  & 40  & 398 \\
% 5  & 328 & 36  & 40  & 404 \\
% 6  & 376 & 42  & 47  & 465 \\
% 7  & 328 & 36  & 41  & 405 \\
% 8  & 181 & 20  & 22  & 223 \\
% 9  & 461 & 51  & 57  & 569 \\
% 10 & 295 & 33  & 37  & 365 \\
% 11 & 228 & 25  & 28  & 281 \\
% 12 & 519 & 58  & 64  & 641 \\
% 13 & 362 & 40  & 45  & 447 \\
% 14 & 364 & 40  & 45  & 449 \\
% 15 & 424 & 47  & 52  & 523 \\
% 16 & 292 & 33  & 36  & 361 \\
% 17 & 435 & 49  & 54  & 538 \\

1 & 420 & 41 & 52 & 513 \\
2 & 151 & 15 & 19 & 185 \\
3 & 525 & 52 & 65 & 642 \\
4 & 326 & 32 & 40 & 398 \\
5 & 330 & 33 & 41 & 404 \\
6 & 380 & 38 & 47 & 465 \\
7 & 331 & 33 & 41 & 405 \\
8 & 182 & 18 & 23 & 223 \\
9 & 466 & 46 & 57 & 569 \\
10 & 298 & 30 & 37 & 365 \\
11 & 229 & 23 & 29 & 281 \\
12 & 522 & 52 & 65 & 639 \\
13 & 366 & 36 & 45 & 447 \\
14 & 368 & 36 & 45 & 449 \\
15 & 428 & 42 & 52 & 522 \\
16 & 295 & 29 & 37 & 361 \\
17 & 440 & 44 & 54 & 538 \\
\hline
\noalign{\vskip 1mm}
\textbf{Total} & \textbf{6057} & \textbf{600} & \textbf{749} & \textbf{7406} \\
\end{tabular}
\end{table}

\subsection{Implementation Details}
\label{subsec:imple-details}

Experiments were conducted on the \textbf{CoAtNet}~\cite{dai2021coatnetmarryingconvolutionattention} model trained on the ImageNet-1k dataset\footnote{\url{https://huggingface.co/timm/coatnet_0_rw_224.sw_in1k}} (referred to as CoAtNet-0) and a more parameter-heavy version pre-trained on the ImageNet-12k dataset, then fine-tuned on the ImageNet-1k dataset\footnote{\url{https://huggingface.co/timm/coatnet_2_rw_224.sw_in12k_ft_in1k}} (referred to as CoAtNet-2). Experiments were also performed on the \textbf{ResNet-50}\footnote{\url{https://docs.pytorch.org/vision/main/models/generated/torchvision.models.resnet50.html}} and \textbf{DenseNet-121}\footnote{\url{https://docs.pytorch.org/vision/main/models/generated/torchvision.models.densenet121.html}} models as baselines for comparison, both initialized with pre-trained weights from their respective models on ImageNet-1k. Additionally, the \textbf{ViT} model\footnote{\url{https://huggingface.co/timm/vit_base_patch16_224.augreg2_in21k_ft_in1k}} was evaluated as a baseline for comparison with a full attention-based model. The number of parameters for each model is presented in Table~\ref{tab:model-params}.

All images were resized to a fixed size of $224 \times 224$ pixels and normalized based on ImageNet channel-wise statistics with mean $\mu = [0.485, 0.456, 0.406]$ and standard deviation $\sigma = [0.229, 0.224, 0.225]$. For data augmentation, horizontal flip combined with MixUp~\cite{zhang2017mixup} and CutMix~\cite{yun2019cutmix} was applied to improve the generalization capacity of the model.

During the model training phase, all models are fine-tuned in an end-to-end manner (i.e., all parameters are fine-tuned) with a batch size of $128$ and were trained for $50$ epochs. The loss function used was cross-entropy, and parameters were optimized using the AdamW~\cite{loshchilov2017decoupled} optimizer with an initial learning rate of $1 \times 10^{-4}$ and weight decay of $1 \times 10^{-3}$. For the learning rate scheduler, cosine annealing with warm restarts was employed, setting $min_{lr} = 0.0$, an initial value for cycle length $T_0 = 3$, and a multiplicative factor $T_{mult} = 1$. The threshold for gradient norm clipping was set to $1.0$ to ensure more stable training. Mixed precision with \textit{fp16} was used for all experiments.

Instead of saving only a single checkpoint for final evaluation, the top $k = 8$ checkpoints were saved for each metric, including \textit{loss}, \textit{accuracy}, and \textit{F1} score, based on the validation set, resulting in a total of $24$ best checkpoints saved (a checkpoint from a single epoch may be saved multiple times). The results are reported based on the checkpoint with the highest \textit{F1} score on the test set; in cases where \textit{F1} scores are equal, \textit{accuracy} is used for comparison (see Appendix~\ref{appendix:checkpoint-selection-explanation} for a more detailed explanation).

\paragraph{Rationale for $k=8$}
We selected $k=8$ checkpoints per metric (loss, accuracy, F1) for \modelsoups construction. This choice reflects a balance between diversity and stability: fewer checkpoints would limit the potential benefit of weight averaging, while too many may introduce noisy or suboptimal snapshots. Empirically, we observed that $k=8$ yields the most stable improvements on the validation and test sets, whereas smaller $k$ led to higher variance and larger $k$ offered no further gains. Moreover, storing eight checkpoints per criterion remains computationally and storage-efficient while providing a sufficiently diverse pool for greedy and uniform soup selection.

\paragraph{Experimental Setup} All experiments were conducted on a system equipped with an AMD EPYC 7702 64-Core Processor (16 physical cores allocated), 31 GB of RAM, and an NVIDIA GeForce RTX 3090 GPU. This configuration provided sufficient computational power for efficient model training and inference.

\subsection{Evaluation Metrics}
\label{subsec:eval-metrics}

To evaluate the performance of our models on the multi-class classification task, the following standard metrics are employed: \textbf{Accuracy}, \textbf{Precision}, \textbf{Recall}, and \textbf{F1-score}. These metrics are computed over the test set and averaged across all classes to account for potential class imbalance (i.e., \textit{macro-averaging}, except for \textbf{Accuracy}, which is calculated by considering the total number of correctly predicted labels across all classes, relative to the total number of test samples).

\begin{itemize}
  \item \textbf{Accuracy} measures the proportion of correctly predicted labels over the total number of test samples:
  \begin{equation}
  \text{Accuracy} = \frac{1}{N} \sum_{i=1}^{N} \mathbb{1}(y_i = \hat{y}_i),
  \end{equation}
  where \( y_i \) is the ground-truth label, \( \hat{y}_i \) is the predicted label, and \( \mathbb{1}(\cdot) \) is the indicator function.

  \item \textbf{Precision}, \textbf{Recall}, and \textbf{F1-score} are computed for each class and then averaged (macro-average) to ensure equal weighting regardless of class frequency:
  \begin{equation}
  \text{Precision}_{\text{macro}} = \frac{1}{C} \sum_{c=1}^{C} \frac{TP_c}{TP_c + FP_c},
  \end{equation}
  \begin{equation}
  \text{Recall}_{\text{macro}} = \frac{1}{C} \sum_{c=1}^{C} \frac{TP_c}{TP_c + FN_c},
  \end{equation}
  \begin{equation}
  \text{F1}_{\text{macro}} = \frac{1}{C} \sum_{c=1}^{C} \frac{2 \cdot TP_c}{2 \cdot TP_c + FP_c + FN_c},
  \end{equation}
  where \( TP_c \), \( FP_c \), and \( FN_c \) denote the number of true positives, false positives, and false negatives for class \(c\), respectively, and \(C\) is the total number of classes.
\end{itemize}

These metrics allow for the assessment of not only overall model performance, but also class-wise behavior, which is critical in applications with class imbalance or semantic diversity.

\subsection{Experiment Results}
\label{subsec:experiment-results}

\subsubsection{Main Results and Analysis}
Table~\ref{tab:main-results} summarizes the classification performance of various baseline models and our proposed CoAtNet-based framework on the ICH-17 test set. Among the conventional architectures, ViT outperforms ResNet-50 and DenseNet-121, achieving an F1-score of 67.41\%, highlighting the potential of attention-based models for ICH classification.

CoAtNet-0 achieves a solid performance with an accuracy of 70.63\% and F1-score of 65.98\%, outperforming all baselines and previous studies without ensembling including ViT, despite having approximately three times fewer parameters (see Table~\ref{tab:model-params}).
This demonstrates the efficiency of CoAtNet’s hybrid design, which combines convolutional and attention mechanisms.

\begin{table}[!h]
    \centering
    \caption{Number of parameters for each model}
    \begin{tabular}{l|c}
        \hline
        \textbf{Model} & \textbf{\texttt{\#} Params (M)} \\ \hline
        ResNet-50      & 23.54                   \\
        DenseNet-121   & 6.97                    \\ 
        ViT            & 85.81                   \\ \hline
        CoAtNet-0      & 26.68                   \\ 
        CoAtNet-2      & 72.86                   \\ \hline
    \end{tabular}
    \label{tab:model-params}
\end{table}

Applying \textit{uniform soup} to CoAtNet-0 leads to improvements in accuracy (+0.67\%), recall (+0.97\%), and F1-score (+0.40\%), though precision decreases slightly. Interestingly, \textit{greedy soup} also improves performance, achieving a slightly higher F1-score (+0.44\%) than uniform soup, with a modest gain in accuracy (+0.40\%) and recall (+0.77\%), albeit with lower precision. These results indicate that even for smaller backbones like CoAtNet-0, weight-space ensembling can be beneficial, though metric trade-offs may vary across strategies.

When scaling up to CoAtNet-2, the performance further improves across all metrics. The base model reaches 71.43\% accuracy and 68.58\% F1-score. Applying uniform soup boosts these values to 72.36\% and 69.28\%, respectively, achieving the best overall results. Greedy soup also yields substantial gains (72.23\% accuracy, 69.05\% F1-score), slightly below uniform soup but consistently better than the base model.

These findings collectively validate the effectiveness of weight-space ensembling via \modelsoups. In particular, the improvements seen with both uniform and greedy soups -- especially in the larger CoAtNet-2 model --  underscore the advantages of checkpoint averaging in enhancing generalization without increasing inference complexity.

\begin{table*}[!ht]
    \centering
    \caption{Classification results on the ICH-17 test set. Asterisk ($^*$) indicates results reported in the original paper, and -- denotes missing results. The values located in the upper right corner of the entries in the rows corresponding to uniform soup and greedy soup indicate the difference compared to the corresponding model without applying model soups.}
    \resizebox{.65\linewidth}{!}{%
    \begin{tabular}{l | c c c c} \toprule
        \textbf{Model} & \textbf{Accuracy} & \textbf{Precision} & \textbf{Recall} & \textbf{F1} \\
        \midrule
        ResNet-50 & 65.55 &	64.72 &	62.33 &	62.80 \\
        DenseNet-121 & 64.35 & 60.33 & 60.33 & 60.02 \\
        ViT & 70.09 & 68.79 & 66.99 & 67.41 \\
        \color{gray}
        Do et al.~\cite{10.1007/978-3-030-38364-0_17} & \color{gray}65.32$^*$ & \color{gray}-- & \color{gray}-- & \color{gray}-- \\
        \color{gray}
        Tran et al.~\cite{tran2024fusing} & \color{gray}66.76$^*$ & \color{gray}-- & \color{gray}-- & \color{gray}-- \\
        \midrule
        CoAtNet-0 & 70.63 & 67.66 & 66.17 & 65.98 \\
        \rowcolor{gray!10}
        CoAtNet-0 + uniform soup & $71.30^{\uparrow 0.67}$ & $66.65^{\downarrow 1.01}$ & $67.14^{\uparrow 0.97}$ & $66.38^{\uparrow 0.40}$ \\
        \rowcolor{gray!10}
        CoAtNet-0 + greedy soup & $\text{71.03}^{\uparrow 0.40}$ & $\text{66.85}^{\downarrow 0.81}$ & $\text{66.94}^{\uparrow 0.77}$ & $\text{66.42}^{\uparrow 0.44}$ \\
        \hdashline
        \noalign{\vskip 1mm}
        CoAtNet-2 & 71.43 & 69.34 & 68.40 & 68.58 \\
        \rowcolor{gray!10}
        CoAtNet-2 + uniform soup & $\textbf{72.36}^{\uparrow 0.93}$ & $\textbf{69.98}^{\uparrow 0.64}$ & $\textbf{69.09}^{\uparrow 0.69}$ & $\textbf{69.28}^{\uparrow 0.70}$ \\
        \rowcolor{gray!10}
        CoAtNet-2 + greedy soup & $\underline{72.23}^{\uparrow 0.80}$ & $\underline{69.91}^{\uparrow 0.57}$ & $\underline{68.81}^{\uparrow 0.41}$ & $\underline{69.05}^{\uparrow 0.47}$ \\
        % \rowcolor{gray!10}
        % \color{gray}CoAtNet-2 + soft voting (Sec. \ref{sec:compare}) & \color{gray}$\text{71.16}$ & \color{gray}$\text{69.06}$ & \color{gray}$\text{67.52}$ & \color{gray}$\text{67.89}$ \\
        \bottomrule
    \end{tabular}
    }%
    \label{tab:main-results}
\end{table*}

\begin{table*}[!ht]
    \caption{Results for each class of the CoAtNet-2 model, where \textit{best} corresponds to the highest result among the saved checkpoints (see~\ref{subsec:imple-details}), and \textit{soups} corresponds to applying a model soup (either uniform or greedy). The \textbf{bolded} values correspond to the highest value in the column for the respective class.}
    \centering
    \resizebox{.65\linewidth}{!}{%
    \begin{tabular}{c | c c | c c | c c | c c} \toprule
        \multirow{2}{*}[0pt]{\textbf{Class}} & \multicolumn{2}{c}{\textbf{Accuracy}} & \multicolumn{2}{c}{\textbf{Precision}} & \multicolumn{2}{c}{\textbf{Recall}} & \multicolumn{2}{c}{\textbf{F1}} \\ \cmidrule(lr){2-3} \cmidrule(lr){4-5} \cmidrule(lr){6-7} \cmidrule(lr){8-9}
        & best & soups & best & soups & best & soups & best & soups \\ \midrule
         1 & \textbf{88.46} & \textbf{88.46} & 80.70 & \textbf{85.19} & \textbf{88.46} & \textbf{88.46} & 84.40 & \textbf{86.79} \\
         2 & \textbf{47.37} & \textbf{47.37} & \textbf{75.00} & 69.23 & \textbf{47.37} & \textbf{47.37} & \textbf{58.06} & 56.25 \\
         3 & \textbf{93.85} & 92.31 & \textbf{95.31} & 95.24 & \textbf{93.85} & 92.31 & \textbf{94.57} & 93.75 \\
         4 & 22.50 & \textbf{25.00} & 28.12 & \textbf{33.33} & 22.50 & \textbf{25.00} & 25.00 & \textbf{28.57} \\
         5 & \textbf{65.85} & \textbf{65.85} & 64.29 & \textbf{71.05} & \textbf{65.85} & \textbf{65.85} & 65.06 & \textbf{68.35} \\
         6 & 61.70 & \textbf{74.47} & 58.00 & \textbf{64.81} & 61.70 & \textbf{74.47} & 59.79 & \textbf{69.31} \\
         7 & 78.05 & \textbf{80.49} & 74.42 & \textbf{76.74} & 78.05 & \textbf{80.49} & 76.19 & \textbf{78.57} \\
         8 & \textbf{43.48} & 34.78 & 40.00 & \textbf{44.44} & \textbf{43.48} & 34.78 & \textbf{41.67} & 39.02 \\
         9 & 50.88 & \textbf{52.63} & \textbf{50.00} & 43.07 & 50.88 & \textbf{52.63} & \textbf{50.43} & 47.24 \\
         10 & 54.05 & \textbf{59.46} & 55.56 & \textbf{57.89} & 54.05 & \textbf{59.46} & 54.79 & \textbf{58.67} \\
         11 & \textbf{86.21} & \textbf{86.21} & \textbf{89.29} & 86.20 & \textbf{86.21} & \textbf{86.21} & \textbf{87.72} & 86.20 \\
         12 & \textbf{87.69} & \textbf{87.69} & 79.17 & \textbf{81.43} & \textbf{87.69} & \textbf{87.69} & 83.21 & \textbf{84.44} \\
         13 & 57.78 & \textbf{64.44} & 68.42 & \textbf{72.50} & 57.78 & \textbf{64.44} & 62.65 & \textbf{68.23} \\
         14 & \textbf{86.67} & \textbf{86.67} & 86.67 & \textbf{95.12} & \textbf{86.67} & \textbf{86.67} & 86.67 & \textbf{90.70} \\
         15 & \textbf{80.77} & \textbf{80.77} & 72.41 & \textbf{75.00} & \textbf{80.77} & \textbf{80.77} & 76.36 & \textbf{77.78} \\
         16 & 64.86 & \textbf{70.27} & \textbf{70.59} & 66.66 & 64.86 & \textbf{70.27} & 67.61 & \textbf{68.42} \\
         17 & \textbf{92.59} & \textbf{92.59} & \textbf{90.91} & \textbf{90.91} & \textbf{92.59} & \textbf{92.59} & \textbf{91.74} & \textbf{91.74} \\
        \bottomrule
    \end{tabular}
    \label{tab:per-class-results}
    }%
\end{table*}

\subsubsection{Comparison between Uniform and Greedy Soup}
To better understand the effects of different weight-space ensembling strategies, we compare \textit{uniform soup} and \textit{greedy soup} applied to the CoAtNet-2 model. As shown in Table~\ref{tab:main-results}, both approaches improve upon the base model across all evaluation metrics, confirming the robustness of \modelsoups in enhancing generalization.

Uniform soup achieves the best overall results, with 72.36\% accuracy and 69.28\% F1-score, outperforming greedy soup by +0.13\% in accuracy and +0.23\% in F1. It also shows slightly higher precision and recall. This suggests that when checkpoint diversity is sufficient, uniform averaging can yield strong and stable performance gains.

Greedy soup, while slightly behind in absolute performance, still improves over the base CoAtNet-2 by +0.80\% in accuracy and +0.47\% in F1-score. Unlike uniform soup, the greedy strategy incrementally selects checkpoints based on validation performance, resulting in a leaner and more selective ensemble. This can be advantageous in scenarios with resource constraints or limited checkpoint quality.

\begin{figure}[!ht]
  \centering
  \includegraphics[width=\linewidth]{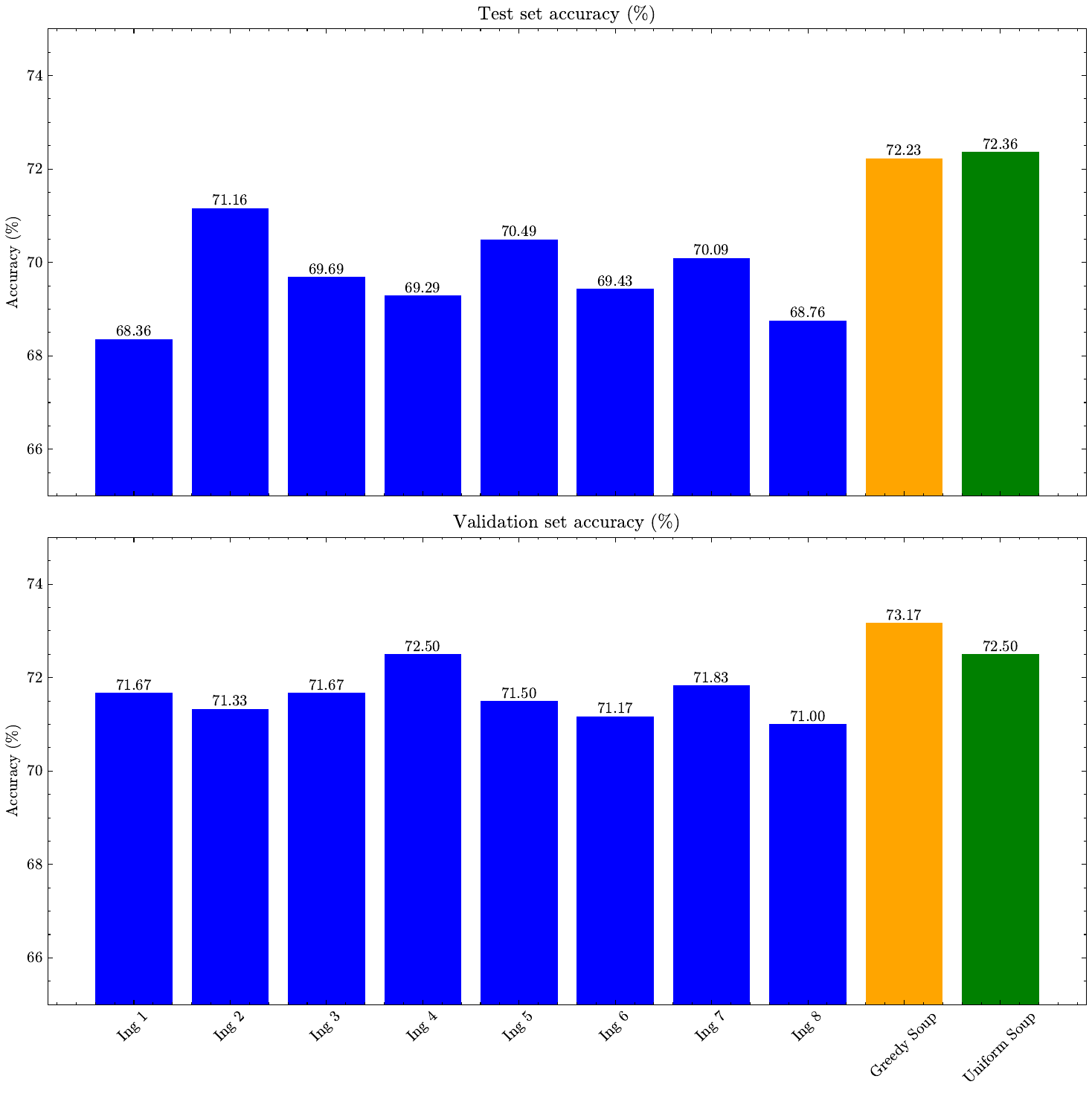}
  \caption{Comparison of validation and test accuracy (\%) between individual CoAtNet-2 models and their combinations via \modelsoups techniques. The ensemble models -- greedy soup and uniform soup -- outperform all individual models, demonstrating the effectiveness of \modelsoups in enhancing generalization}
  \label{fig:ingredient-models}
\end{figure}

Figure \ref{fig:ingredient-models} illustrates the accuracy of the individual models (CoAtNet-2) and the ensemble models generated via \modelsoups techniques, including greedy soup and uniform soup. It is evident that the individual CoAtNet-2 models consistently yield lower accuracy compared to the \modelsoups ensembles.

Interestingly, on the smaller CoAtNet-0 model, the performance gap narrows: uniform soup gains +0.40\% in F1-score, while greedy soup yields a slightly higher improvement of +0.44\%, despite lower precision in both cases. This variation highlights that the effectiveness of each ensembling strategy may depend on model capacity and the diversity of saved checkpoints.

Overall, while uniform soup offers the best results in our setting, greedy soup remains a competitive alternative, especially when computational constraints or checkpoint sparsity are present.
% The observed behavior also reflects the typical limitations of greedy algorithms, as discussed in Appendix~\ref{appendix:coin-change}. 

\subsubsection{Per-Class Performance Analysis}
To further understand the behavior of our model, we conduct a per-class analysis of CoAtNet-2 in both its best individual checkpoint and the \modelsoups variant, as shown in Table~\ref{tab:per-class-results}. The results indicate that \modelsoups yields performance improvements in the majority of classes across all evaluation metrics.

Notably, \modelsoups enhances classification accuracy in 7 out of 17 classes and improves F1-score in 11 classes. For example, class 6 (\textit{Lễ hội Ok Om Bok của người Khmer}) sees accuracy rise from 61.70\% to 74.47\% and F1-score from 59.79\% to 69.31\%, suggesting that ensemble averaging stabilizes predictions in complex or ambiguous categories. Similar improvements are observed in classes 4 (\textit{Lễ hội cúng biển Mỹ Long}), 7 (\textit{Lễ hội Vía Bà Chúa Xứ núi Sam}), 10 (\textit{Lễ làm chay}), and 13 (\textit{Lễ cúng Việc}), which benefit from consistent gains in both precision and recall.

In some cases, performance remains unchanged (e.g., classes 1, 2, 5, 11, 12, 14, 15, and 17), or shows slight drops in certain metrics. For instance, class 8 (\textit{Đại lễ Kỳ yên Đình Tân Phước Tây}) exhibits a drop in accuracy and recall, though precision slightly increases. Similarly, class 16 (\textit{Lễ hội anh hùng Trương}) shows a modest increase in accuracy but a slight decrease in precision. These variations may stem from overfitting in some checkpoints or limited representation during soup selection.

Overall, in terms of F1-score, \modelsoups improves or maintains performance in 12 out of 17 classes, reinforcing its robustness and reliability across diverse cultural heritage categories with varying visual complexity and sample sizes. This per-class evaluation confirms that weight-space ensembling not only improves average-case metrics but also contributes to more stable and generalized class-wise performance.

% \newpage
\section{Analytically Comparing Model Soups to Soft Voting}
\label{sec:compare}

Given a dataset $\mathcal{D} = \{x_1, x_2, \dots, x_N\}$ of $N$ samples, each sample is passed through a model $f$, followed by a softmax layer, resulting in a probability vector $\mathbf{p}^{(f)}_i \in \Delta^{C-1}$ for each input $x_i$, where $\Delta^{C-1}$ denotes the $(C{-}1)$-dimensional probability simplex. Collectively, the model $f$ produces a matrix of softmax outputs:

\[
\mathbf{P}^{(f)} = 
\begin{bmatrix}
(\mathbf{p}^{(f)}_1)^T \\
(\mathbf{p}^{(f)}_2)^T \\
\vdots \\
(\mathbf{p}^{(f)}_N)^T
\end{bmatrix}
\in \mathbb{R}^{N \times C}
\]

To compare two models $f$ and $g$, the mean cross-entropy between their corresponding softmax outputs across the dataset is computed:

\begin{equation}
\label{eq:symmetric_dist}
\begin{aligned}
\text{Dist}(f, g) &= \frac{1}{N} \sum_{i=1}^{N} H_{\text{sym}}\left(\mathbf{p}^{(f)}_i, \mathbf{p}^{(g)}_i\right), \\
\text{where} \quad H_{\text{sym}}(\mathbf{p}, \mathbf{q}) &= \left[ H(\mathbf{p}, \mathbf{q}) + H(\mathbf{q}, \mathbf{p}) \right], \\
H(\mathbf{p}, \mathbf{q}) &= -\sum_{j=1}^{C} p_j \log q_j
\end{aligned}
\end{equation}

This yields a symmetric distance matrix $D \in \mathbb{R}^{|\mathcal{F}| \times |\mathcal{F}|}$ over a set of models $\mathcal{F}$. To visualize the relational structure between models, \textit{Multidimensional Scaling (MDS)}~\cite{demaine2021multidimensional} is applied to project the distance matrix $D$ into a two-dimensional space. MDS is particularly well-suited for this task because pairwise distances are preserved in the low-dimensional embedding, enabling the relative similarity between models to be captured and interpreted based purely on their output behaviors. Unlike other nonlinear techniques such as t-SNE~\cite{cai2022theoreticalfoundationstsnevisualizing}, which may distort global geometry, both local and global distances are maintained by MDS, rendering it more appropriate for preserving structural fidelity in model comparison tasks.

\subsection*{Model Soups Diversity Justification}

The aforementioned pairwise distance matrix $D$ and its corresponding MDS projection are also utilized as a basis for analyzing model diversity in the context of \modelsoups. Specifically, the 2D embedding space is employed to visualize the distribution of candidate models and to evaluate whether the models selected for ensembling (i.e., those included in the soup) are well-spread in output space.

Let $\mathcal{S} \subset \mathcal{F}$ denote the subset of models chosen for \modelsoups. If \modelsoups is effective at capturing a diverse ensemble, the points corresponding to $\mathcal{S}$ in the MDS embedding are expected to exhibit broad coverage of the model space -- avoiding clustering and instead spanning distinct regions.

In the experiments conducted, the selected models $\mathcal{S}$ are highlighted in the MDS scatter plot and observed to be well-separated, confirming that the soup does not merely average redundant models, but rather aggregates models with heterogeneous prediction behaviors. This diversity in softmax outputs contributes to the ensemble’s robustness and improved generalization.

Thus, the analysis provides empirical evidence that the \modelsoups construction process implicitly favors diverse members in the model space, a property known to be beneficial for ensembling in both theory and practice.

\subsection*{Visualizing Model Diversity on the Validation Set}

\begin{figure}[ht]
  \centering
  \includegraphics[width=\linewidth]{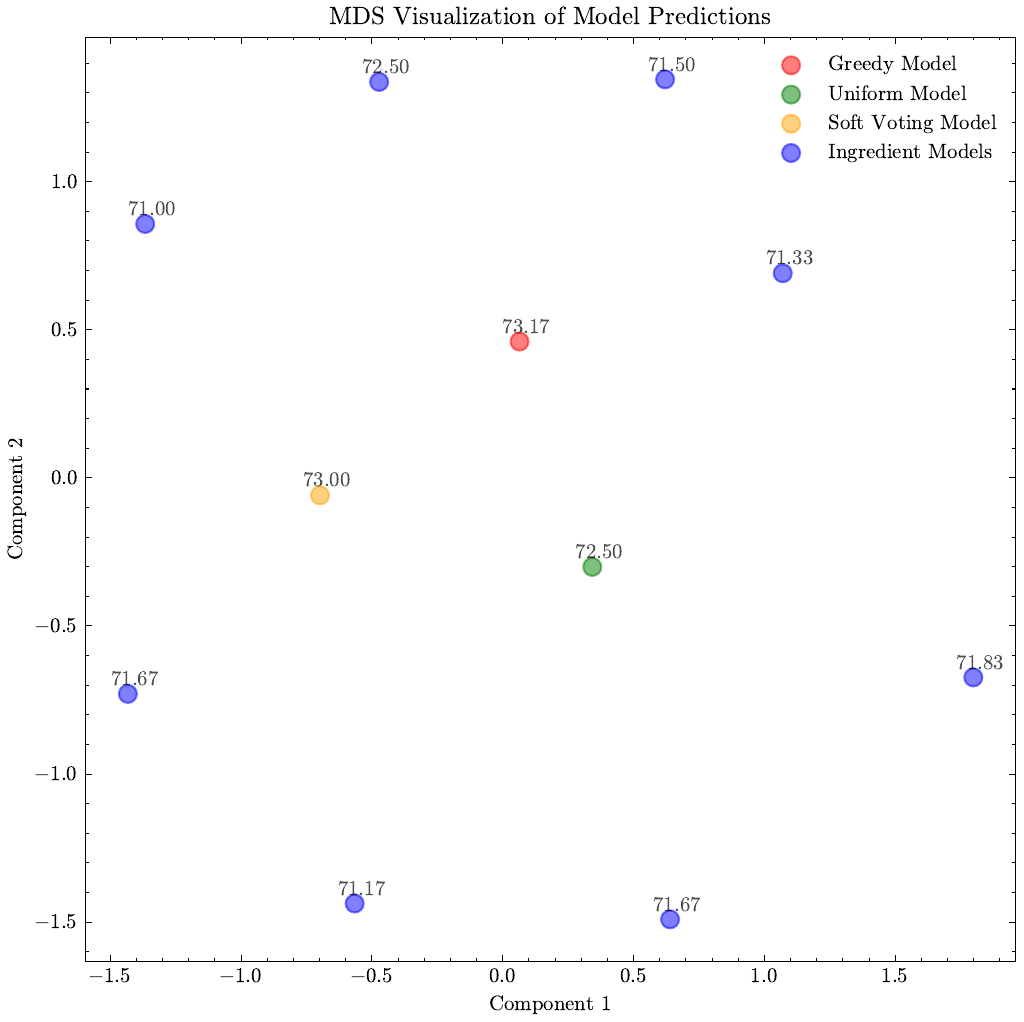}
  \caption{
        \textbf{MDS visualization of model predictions on the validation set.} Each point represents a model projected via MDS using cross-entropy-based pairwise distances. The ingredient models (blue) are broadly distributed, while ensemble models -- Greedy (red), Uniform (green), and Soft Voting (orange) -- cluster near the center. This spatial structure confirms that \modelsoups leverages predictive diversity more effectively than Soft Voting. The value above each point represents the \textit{accuracy} on the validation set for the corresponding model.
    }
  \label{fig:val_visualize_pick}
\end{figure}

To evaluate the diversity of models selected by \modelsoups, the softmax outputs (i.e., probabilities) of all candidate and ensemble models are projected using Multidimensional Scaling (MDS) based on pairwise cross-entropy distances. The validation set is chosen for visualization (Figure~\ref{fig:val_visualize_pick}), as it reflects the basis on which checkpoint selection is performed by both greedy and uniform soup strategies.

The resulting 2D embedding reveals that the ingredient models (blue) are well-dispersed across the output space, indicating a high degree of predictive diversity. In contrast, the ensemble models -- Greedy (red), Uniform (green), and Soft Voting (orange) -- are observed to be clustered near the center of this distribution. This geometry suggests that \modelsoups draws from a broad range of behaviors and aggregates diverse checkpoints, while Soft Voting averages across more redundant predictions.

The geometric separation among the ingredients confirms that \modelsoups does not merely average similar models but actively selects complementary ones. Such diversity is critical for reducing variance in ensemble predictions and improving generalization. This visualization provides compelling evidence that \modelsoups functions as a diversity-aware selection strategy, offering both theoretical and empirical benefits over traditional voting-based ensembles.

% \subsection*{Comparison with Soft Voting}

% It is important to contrast the model soup strategy with the classical Soft Voting ensemble. In Soft Voting, all candidate models contribute equally to the final prediction, regardless of their individual performance or diversity. This uniform averaging often leads to ensembling over redundant models, which may cluster closely in the MDS space, as seen in Figures~\ref{fig:train_visualize_pick}--\ref{fig:test_visualize_pick}. In our visualizations, the Soft Voting model (orange) consistently appears near the center of the ingredient cloud, suggesting that it blends overlapping predictive behaviors rather than exploiting complementary ones.

% By contrast, model soup (particularly in its greedy or uniform selection variants) explicitly selects a subset of models to optimize validation performance. Our empirical analysis shows that these selected models are not only individually strong but also distributed across the output space — a hallmark of diversity. This targeted selection strategy enables model soup to outperform Soft Voting by avoiding over-representation of similar models and instead leveraging heterogeneity to achieve superior ensemble generalization.

% Thus, our results highlight a key strength of model soup: it serves not merely as a voting mechanism, but as a diversity-aware model selection strategy that dynamically adapts to the geometry of the prediction space.

In Soft Voting, the ensemble prediction is computed by averaging the output probability vectors from all individual models in the ensemble. Given a set of $M$ trained models $\{f^{(1)}, f^{(2)}, \dots, f^{(M)}\}$, the ensemble output for an input $x$ is defined as:

\[
\mathbf{p}^{\text{soft}}(x) = \frac{1}{M} \sum_{m=1}^{M} \mathbf{p}^{(f^{(m)})}(x)
\]

where $\mathbf{p}^{(f^{(m)})}(x) \in \Delta^{C-1}$ denotes the softmax output of model $f^{(m)}$ for input $x$, and $\Delta^{C-1}$ is the $(C{-}1)$-dimensional probability simplex over $C$ classes.

The final predicted class is then:

\[
\hat{y} = \arg\max_{j} \left[ \mathbf{p}^{\text{soft}}(x) \right]_j
\]

This method treats all models as equally informative, regardless of their individual accuracy or diversity, which may limit its effectiveness when model predictions are highly correlated.

\subsection*{Comparison with Soft Voting and Bias-Variance Analysis}

To better understand the role of \modelsoups in improving generalization, it is compared to the classical Soft Voting ensemble. In Soft Voting, all models contribute equally to the final prediction through uniform averaging of their softmax outputs, regardless of their individual performance or redundancy. This approach does not explicitly account for diversity among models, and can lead to ensembling over highly similar checkpoints. As shown in Figure~\ref{fig:val_visualize_pick}, the Soft Voting ensemble (orange) is located near the center of the output-space embedding, reflecting its averaging over clustered, potentially redundant models.

In contrast, \modelsoups\ -- particularly in its \textit{greedy} and \textit{uniform} forms -- constructs the ensemble by selectively averaging a subset of validation-optimal checkpoints that are geometrically diverse in the output space. This selective diversity is not incidental; from a bias–variance perspective, averaging over complementary models with uncorrelated errors is known to reduce prediction variance while introducing minimal additional bias. The analysis demonstrates that \modelsoups leverages this property effectively, functioning as a variance-reducing, diversity-aware ensemble strategy.

Overall, \modelsoups goes beyond naive output averaging: it dynamically adapts to the geometry of model outputs and the validation objective, yielding more robust predictions under low-resource and high-variance conditions compared to Soft Voting.

\begin{table*}[h]
\centering
\caption{Comparison of Model Soups and Soft Voting on key ensemble properties.}
\label{tab:soup_vs_softvoting}
\resizebox{.8\linewidth}{!}{%
\begin{tabular}{l|c|c}
\hline
\textbf{Aspect} & \textbf{Soft Voting} & \textbf{Model Soups} \\
\hline
\textbf{Inference Cost} & High (all models active) & Low (single averaged model) \\
\textbf{Memory Footprint} & Scales with number of models & Constant (single model) \\
\textbf{Deployment Simplicity} & Requires parallel model storage & Simple (one model checkpoint) \\
\textbf{Bias Reduction} & Limited & Slight increase (depends on soup strategy) \\
\textbf{Variance Reduction} & Moderate (correlated models) & High (diverse checkpoint averaging) \\
\hline
\end{tabular}
}%
\end{table*}

\section{Ablation Study: Effect of Removing Pre-training}

To further investigate the impact of pre-trained initialization, an ablation study was conducted where both CoAtNet-0 and CoAtNet-2 were trained from random initialization rather than being fine-tuned from ImageNet pre-trained weights. The models were trained under the same experimental settings as in the main experiments, and their performance was evaluated on the ICH-17 test set.

\begin{table*}[h]
\centering
\caption{Performance of CoAtNet models trained from random initialization on the ICH-17 test set. The values located in the upper right corner of the entries in the rows corresponding to uniform soup and greedy soup indicate the difference compared to the corresponding model without applying \modelsoups.}
\label{tab:ablation-random}
\resizebox{.65\linewidth}{!}{%
\begin{tabular}{lcccc}
\toprule
Model & Accuracy & Precision & Recall & F1 \\
\midrule
CoAtNet-0 & 49.13 & 46.36 & 45.31 & 44.85 \\
% CoAtNet-0 + uniform & 1 & 2 & 3 & 4 \\
% CoAtNet-0 + greedy & 1 & 2 & 3 & 4 \\
CoAtNet-2 & 51.27 & 48.87 & 48.21 & 47.81 \\
\hdashline
\noalign{\vskip 1mm}    
CoAtNet-2 + uniform & $52.07^{\uparrow 0.80}$ & $49.27^{\uparrow 0.40}$ & $48.82^{\uparrow 0.61}$ & $47.99^{\uparrow 0.18}$ \\
CoAtNet-2 + greedy & $51.80^{\uparrow 0.53}$ & $48.72^{\downarrow 0.15}$ & $48.77^{\uparrow 0.56}$ & $48.11^{\uparrow 0.30}$ \\
\bottomrule
\end{tabular}
}%
\end{table*}

As shown in Table~\ref{tab:ablation-random}, both models trained from scratch perform significantly worse compared to their ImageNet-pretrained counterparts (CoAtNet-0: 70.63\% accuracy, 65.98\% F1; CoAtNet-2: 71.43\% accuracy, 68.58\% F1). Training without pre-training results in drops of approximately 20--22 percentage points in accuracy and 21--23 percentage points in macro F1-score. These results highlight that pre-training provides critical inductive biases and transferable representations, which are especially important in the low-resource and noisy ICH-17 dataset.

\section{Conclusion and Future Work}
\label{sec:conclusion}

In this study, we introduced a robust framework for classifying images of Intangible Cultural Heritage (ICH) in the Mekong Delta by combining the hybrid CoAtNet architecture with weight-space ensembling via \modelsoups. By leveraging the representational power of CoAtNet and the generalization benefits of checkpoint averaging, our method achieves state-of-the-art performance on the ICH-17 dataset. We adopted both uniform and greedy soup strategies, outperforming strong baselines such as ResNet-50, DenseNet-121, and ViT across global accuracy and per-class F1-score metrics.

Beyond performance gains, a comprehensive analysis of model diversity was conducted using cross-entropy-based distances and Multidimensional Scaling (MDS). This analysis reveals that the models selected by the soup strategy span diverse regions in output space -- unlike classical Soft Voting, which averages over tightly clustered models. The findings empirically confirm that \modelsoups acts not merely as an averaging scheme, but as a diversity-aware selection mechanism that enhances robustness and generalization across training, validation, and test splits.

For future work, this framework is planned to be extended in two directions. First, integrating semantic priors and multi-modal signals (e.g., textual metadata) will be investigated to further enhance performance in low-resource cultural datasets. Second, the methodology is intended to be scaled to broader ICH datasets across other regions, contributing to inclusive and AI-driven preservation of cultural heritage at both regional and global levels.

\section*{Acknowledgment}
We would like to express our sincere gratitude to the AniAge project for providing the ICH dataset used in our experiments. This valuable resource enabled a fair and rigorous evaluation of the proposed methods in a culturally grounded context.

\section*{Data Availability}

The dataset used in this study is not publicly available due to institutional or licensing restrictions. However, it can be made available for academic use upon reasonable request. Interested researchers may contact the authors for further information.

\bibliographystyle{splncs04}
\bibliography{refs}

\vfill

\begin{appendices}

\section{The Method for Selecting the Best Checkpoint in Our Approach}
\label{appendix:checkpoint-selection-explanation}

Instead of selecting only a single checkpoint, such as the one with the highest accuracy or lowest loss based on the validation set, the top $k = 8$ checkpoints for each metric, including loss, accuracy, and F1 score, were saved based on the validation set, resulting in a total of $24$ best checkpoints saved. These checkpoints were then evaluated on the test set, and the results of the checkpoint with the highest F1 score were reported. In cases where multiple checkpoints had equal F1 scores, they were compared based on accuracy. This approach was adopted because, during the experimental phase, it was observed that some results on the validation set were significantly high, yet a relatively large gap was found on the test set (in some cases, up to a 4\% difference in accuracy). This could lead to a scenario where the checkpoint with the highest validation performance is selected, resulting in relatively poor test set performance.

We hypothesize that this phenomenon is due to the characteristics of the ICH-17 dataset, where the context of the images is highly diverse, and the data remains relatively noisy, reducing the model’s generalization capability. Furthermore, the number of images per class in both the test and validation sets is limited when considered individually, combined with the diverse contexts of the images, which may lead to an inconsistent distribution between the validation and test sets.

However, it should be noted that this does not affect the application of \modelsoups, as evaluation and checkpoint selection were performed entirely based on the validation set results, and the test set was only evaluated on the final checkpoint aggregated from the uniform soup and greedy soup algorithms.

\end{appendices}

\end{document}